\documentclass{svproc}
\usepackage[numbers,sort&compress]{natbib}
\usepackage{url}

\usepackage{graphicx,xcolor}
\usepackage{tabularx}
\usepackage{soul}
\usepackage{subcaption}
\usepackage{booktabs}

\usepackage{epstopdf}
\usepackage[latin1]{inputenc}

\usepackage[hidelinks]{hyperref}
\usepackage{xstring}

\begin{document}
\mainmatter              
\title{Bridging the Domain Gap for Stance Detection for the Zulu language}
\titlerunning{Bridging Domains for Zulu Stance Detection}  
%

\author{Gcinizwe Dlamini\inst{1} \and Imad Eddine Ibrahim BEKKOUCH\inst{2} \and Adil Khan\inst{1} \and Leon Derczynski\inst{3}}

\authorrunning{G Dlamini et al.}

\institute{Innopolis University, Tatarstan, Russian Federation,\\
\and  Sorbonne Center for Artificial Intelligence, Sorbonne University, Paris, France
\and IT University of Copenhagen, Denmark}

\maketitle              

\begin{abstract}
Misinformation has become a major concern in recent last years given its spread across our information sources. In the past years, many NLP tasks  have been introduced in this area, with some systems reaching good results on English language datasets. Existing AI based approaches for fighting misinformation in literature suggest automatic stance detection as an integral first step to success. Our paper aims at utilizing this progress made for English to transfers that knowledge into other languages, which is a non-trivial task due to the domain gap between English and the target languages. We propose a black-box non-intrusive method that utilizes techniques from Domain Adaptation to reduce the domain gap, without requiring any human expertise in the target language, by leveraging low-quality data in both a supervised and unsupervised manner. This allows us to rapidly achieve similar results for stance detection for the Zulu language, the target language in this work, as are found for English. We also provide a stance detection dataset in the Zulu language. Our experimental results show that by leveraging English datasets and machine translation we can increase performances on both English data along with other languages.
\end{abstract}

\section{Introduction}

Social media platforms have become a major source of information and news. 
At the same time, the amount of misinformation on them has also become a concern. 
Automatic misinformation detection is a challenging task. 
One was of categorising rumours is into those that can be grounded and verified against a knowledge base, and those that cannot (e.g. due to a lack of knowledge base coverage). 
This task has required the work of professional fact-checkers. 
This work has recently been complemented by fact-checking systems. 
However, fact checking only works when there is evidence to refute or confirm a claim.
Automated fact checking relies on databases for this evidence.
These databases often lack the information needed, due to e.g. lag or a lack of notability.
In this case, another information source is needed.
Prior work has hypothesised~\cite{qazvinian2011rumor} and shown that the stance, that is the attitude, that people take towards claims can act as an effective proxy for veracity predication~\cite{dungs2018can,lillie2019joint}.

A primitive solution to analyzing texts written in uncommon languages, i.e. those  which are not supported by the state-of-the-art NLP models, 
is to translate the text to a language an appropriate NLP model has been trained on. However, this solution is not always effective and does not give good performance. Some of the main difficulties faced by state-of-the-art NLP models (especially in the task of stance detection using translated LREL) come from: discrepancies induced by the LREL translation process; LREL data scarcity \cite{allah2012toward}; and the noise that exists in the social media data itself~\cite{derczynski2013microblog}.

An alternative approach to solving this problem can be ensemble models for stance detection \cite{KOTU201919} trained on different languages and combining their respective results, but we still face the same problem which is the fact that each model is built from a well-resourced, well-represented language, so if we introduce another language to it even by translation, its individual performance is not likely to be good, hence the overall ensemble performance will probably be the same as the previously suggested solution \cite{NISBET2009285}.

Thus, in light of the problem of wide spread of false rumours, mis-information, and challenges faced by state-of the-art NLP models when it comes to LREL, we propose an approach to solving these challenges in this paper. We propose using Domain Adaptation (DA)\cite{DBLP:journals/corr/KochkinaLA17} in the task of Stance Detection on LREL Twitter text data.


In our approach we use machine translation to translate the model's training data, taking into account the amount of noise this introduces. Our intended stance detection model will be able to learn from variety of low quality data, whereby the low quality aspect comes from the automated translation. This comes up as a strength in our approach for two reasons: 1. For most social media text classification tasks, precise and canonical forms of expression are not the most important feature for making the decisions. 2. The target data (The LREL) has an element of being noisy since it's from social media and is a direct output of a translator. Our method uses data from multiple languages to build a stance detection model for a less-privileged language. 

In addition to the proposed Domain Adaptation technique for stance detection, we contribute a new dataset for Stance Detection in a language previously without resources. 

The structure of this paper is as follows: Section 2 outlines the related works. In Section 3, the model architecture, methodology and datasets used in detail, while Section 4 presents the obtained results followed by discussion. In conclusion Section 5 we discuss possible future directions and insights from the paper results.

\section{Related Work}
Researchers have approached this problem of fake news detection, mis-information and stance detection
from many point of views. Our approach is about combining concepts from different areas of research, mainly: Domain Adaptation \cite{Bekkouch_2019}, Domain Generalization\cite{DBLP:journals/corr/abs-1710-03077}, Domain Randomization \cite{weng2019DR} and some old and new techniques where the concept of using data from one language to improve performances on another was used or from one task to improve performances on another \cite{mahsut2001utilizing}. 

\subsection{Domain Generalization,  Adaptation,  Randomization}

These three fields of research are very tightly related, the basic idea behind them is to use multiple simple to collect data sources to improve performance on a harder to collect (or label) dataset. Domain Adaptation(DA)\cite{Bekkouch_2019} is the most widely researched topic and specifically Unsupervised Domain Adaptation (UDA) where we use both data and labels from one source domain and use only the data of a target domain without its labels and try to build a model that gives good performances on both domains. The most common method for performing UDA is by utilizing Generative Adversarial Networks\cite{goodfellow2016nips} in multiple ways \cite{8578260} , but there are other techniques that use simpler and faster techniques based only on adversarial losses. 

Domain Generalization on the other hand uses multiple source domains and aims at generalizing to unseen target domains. The current state if the art in DG is leaning towards improving the ability to learn and merge knowledge both from supervised and unsupervised learning. the supervised part is by classifying samples into their corresponding labels whereas the unsupervised part leverages only data in many ways, one way is by reorganising the images into a jigsaw puzzle and training a classifier to solve this\cite{DBLP:journals/corr/abs-1903-06864}. 

Domain Randomization \cite{DBLP:journals/corr/TobinFRSZA17} is the extreme case where we only have access to one domain on training and we want to improve the results on unseen domains. Most of these techniques agree on the fact that having a lot of messy and non-perfect data that comes from multiple sources improves accuracy and allows the model to perform better on real data\cite{weng2019DR}.

\subsection{Stance Detection}
One of the core approaches to automatic fake news assessment is stance detection which is the extraction of a subject's reaction to a claim made by a primary actor \cite{DBLP:journals/corr/AugensteinRVB16}. The main approach taken by NLP researchers for the past years has been shifted towards less hand engineered techniques and into using Deep Learning by taking two text sequences, encoding them in some form (mainly by adding a mask that determines which word belongs to which sentence), and then estimating the type of relationship that joins them. 

One way of formulating stance detection was to divide it into two sub-tasks \cite{mohammad2016semeval}, Task A: fully supervised, classify labeled texts into "IN-FAVOR", "AGAINST", "NONE". The text belongs to 5 topics: "Atheism", "Climate Change is a Real Concern", "Feminist Movement", "Hillary Clinton", and "Legalization of Abortion". Task B on the other hand aims at solving weakly supervised stance detection by training on data from Task A and unlabeled data from another topic and measuring the performance on it. The goal of this paper is actually a next step on top of these two Tasks which is sharing the knowledge from labeled data into another dataset that comes from a totally different language not just a different topic. Zhou et al.\cite{8851965} introduced an approach based on convolution neural networks(CNN) and Attention to detect stances in tweeter. Their proposed approach addresses the challenges faced by CNN's which is generating and capturing high quality word embedding having global textual information. Less Resourced and Endangered Languages are not addressed in their proposed approach for stance detection.

Other major efforts in stance detection include the RumourEval challenge series~\cite{gorrell2019semeval}. Work on LRELs includes datasets in Czech~\cite{hercig2018stance} and Turkish~\cite{kuccuk2017stance}.

\subsection{Explicit \& Implicit Transfer Learning for NLP}
Adapting Computational Language Processing (CLP) \cite{ferraro2013improving} techniques were used in the early 2000s as a first approach to transfer knowledge from one language to the other which we can see in the 'MAJO system' which uses the similarities between Japanese and Uighur to improve substantially the performance\cite{mahsut2001utilizing}. This idea although non-generic and a completely human-centric approach it is one of the first approaches of improving results on a LREL by using a more popular and widely used language. 

Another more recent approach is Bidirectional Encoder Representations from Transformers (BERT)\cite{devlin2018BERT}. BERT is built on the idea of building a general purpose model for NLP that is trained on large amounts of text data and can be easily fine tuned to downstream NLP tasks like stance detection for example.
BERT also has a multilingual mode that has learned on data from 104 languages.

Another very popular approach to Inductive transfer learning in NLP is Universal Language Model Fine-tuning (ULMFiT)\cite{DBLP:journals/corr/abs-1801-06146}, which aims at reducing the amount of labeled data needed for building any NLP tasks and specially for classification. It works on leveraging the the existence of big amounts of unsupervised text that can be used to train a general purpose model which can later be fine-tuned in two steps, the first one is unsupervised where the model learns from raw text that is related to the problem and then the second step is supervised where we train the final layers of the model using a small amount of labeled data, for example we train a general model on Wikipedia texts and then fine-tune it for emails by using a large amount of unlabeled emails in the first step and just a hand-full of labeled emails in the second. 

From all the existing state-of-the-art approaches in stance detection, breakthroughs of transfer learning and domain generalization great performance in lot of different domains, we have found that there still exists a need to cover LRELs in the domain of NLP, specifically in stance detection. With our approach we are hoping to contribute to the existing methods and help the research community develop methods for fighting mis-information, fake news detection and understanding social media content better as compared to existing state-of-the-art methods.

\section{Architecture, Methodology And Dataset}
Our method consists of a two step process. Firstly, the creation of the training Dataset where training dataset is retrieved and joined from multiple languages in order to train a more resilient stance detection model. Secondly, building the training pipeline whereby the chronology for implementing stance detection for less-privileged languages is outlined.

We use Zulu as a case study for a less-privileged language.
The Zulu language~\cite{cope1966Zulu} is a good example of a Less Resourced and Endangered Language (LREL)~\cite{besacier2014automatic} and uncommon language found in social media. Zulu is spoken by an estimated 10 million speakers, mainly located in the province of KwaZulu-Natal of South Africa and Swaziland. Zulu is recognised as a one of the 11 official South African languages and it is understood by over 50\% of its population. As interesting as it is, there are not enough Zulu language resources on the internet to help build major NLP models; and even though Google has done an amazing job with providing models that can be fine-tuned for specific tasks on more than 104 languages using BERT-multilingual \cite{DBLP:journals/corr/abs-1906-01502}, Zulu and many other languages were not part of their research.

\subsection{Step 1: Build the training Dataset} 
Our aim is to have a dataset large enough to expose an underlying structure of a given topic in LREL for any NLP model to capture and generalize.
For this reason we gather our dataset from multiple sources which are chosen randomly from the languages supported by google translate. The total number of sources ranges from 2 to 106.

More formally, we will have a training dataset coming from $N$ languages, where $N_i$ is the number of labeled samples in the $i$th source dataset, such that $$ X^s_i = \{(x_{i,j}^s,y_{i,j}^s)\}_{j=1}^{N_i}$$ where $x_{is}^j$ denotes the $jth$ text sample from the $ith$ source dataset and $y_{is}^j$ is its label.

In the case where there is just a source dataset $ X^s_i$ that gets translated into other languages we denote the $kth$ translated version as  $X^s_{ik}$.

We denote also $M$ the number of labeled samples in the target dataset and $$ X^t = \{(x_{j}^t,y_{j}^t)\}_{j=1}^{M}$$ where $x_{j}^t$ denotes the $jth$ text sample from the target dataset and $y_{j}^t$ is its label. 

In adoption of the ensemble learning principle of using different training data sets \cite{KOTU201517}, we select data sources with different marginal distributions (i.e. they have some dissimilarities between them in terms of syntax rules, language family tree, e.t.c) and from the dataset of stance detection.
These differences in marginal distributions are at the core of the strength of our stance detection model 
since the input of the model will be a translator's result we will force the model to actually learn from the translator results. This idea can also be used in the case where your input data isn't perfect; for example if your model is being deployed on a task where the target audience isn't a language expert they will most likely make many mistakes in their writing, your model should be familiar with these mistakes and more resilient to them. Such mistakes can be easily generated from translators where we can translate from language A to language B and then vice versa, giving us more variety in the data which can be considered as a data augmentation technique. 

\begin{figure}[!ht]
\centering
  \begin{subfigure}{\textwidth}
    \centering\includegraphics[width=\textwidth]{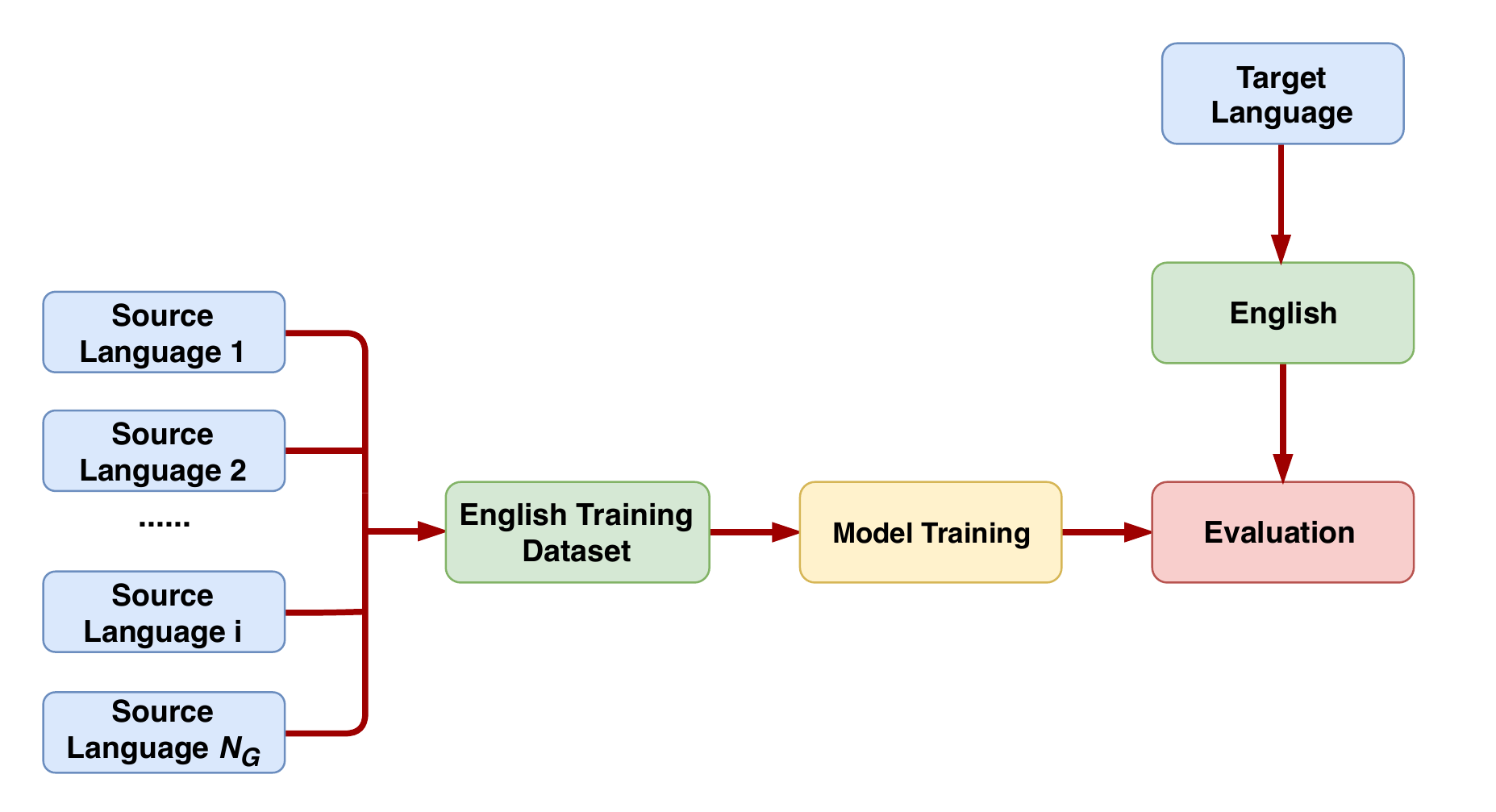}
    \caption{Domain Generalization: multiple source languages all converted into English to use pre-trained models.}
    \label{figure: before_DA}
  \end{subfigure}
  \begin{subfigure}{\textwidth}
    \centering\includegraphics[width=\textwidth]{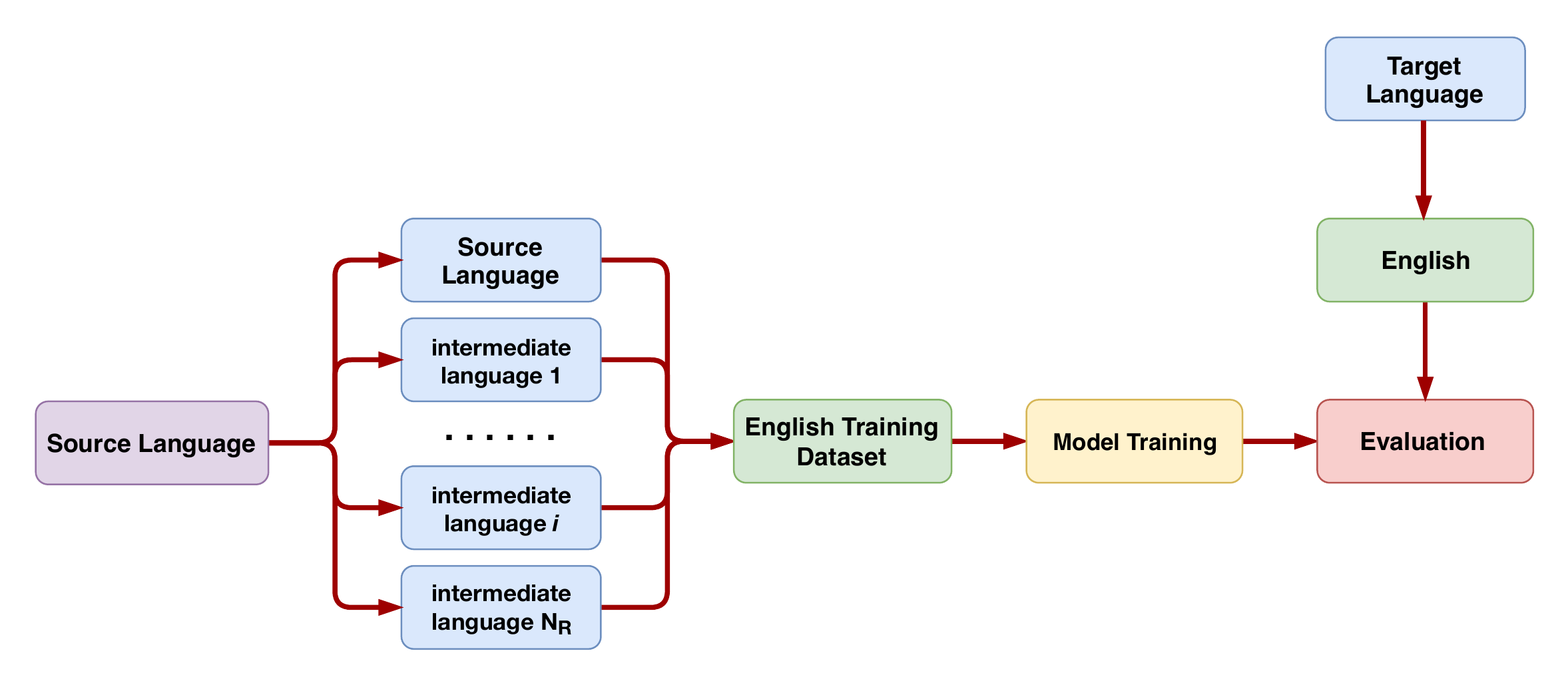}
    \caption{Domain Randomization: one source language to which we add multiple translated versions as data augmentation.}
    \label{figure: after_DA}
  \end{subfigure}
  \caption{Dataset construction process for Domain Generalization and Domain Randomization.}
  \label{figure:data_construct}
\end{figure}

Modeling this domain gap in the training data can be done in one of two ways which are shown on Figure\ref{figure:data_construct}. The domain gap in Text datasets can be anything from the length of sentences, to sentiment changes and even difference in the grammatical correctness of the sentences. In our case we are mostly focused on modeling the errors that a translator might make, and putting them in our training data so that we can correctly classify them on inference time.

\begin{description}
\item[Domain Generalization:] For DG we  have available for us multiple datasets similar in structure and purpose but come from different languages, we translate them all to English and build our classifier. We denote $n_G$ as the number of datasets used to build the model. 

Although there are no constraints on which family the language comes from, but empirical results show that using datasets from different families help to generalize better to unseen datasets and using datasets from the same family as the target language helps for generalizing to it better.

\item[Domain Randomization:] In this case we only have one dataset from one language , so we apply some randomization to increase the size of the dataset and make it more inclusive to mistakes; We do that by translating the dataset into multiple intermediate languages and then translate them into English. The results empirically appear to contain many mistakes but the huge increase in size makes up for it. We denote $n_R$ as the number of intermediate languages used to build the model.

Same remarks as for DG, using intermediate languages that are quite different allows the model to learn from an even richer dataset and overall generalizes better to unseen domains, whereas for a specific target domain, using languages that are from it's same family allows the model to perform better on this target dataset.
\end{description}

\subsection{Step 2: Build the training  pipeline}
Now that the data set is ready, the next step will be to build a pipeline that inputs the dataset, cleans it, tokenizes it, convert it into word embeddings and train the model on it. This step is the reason why we convert all of our datasets into English rather than the target language it self or some other language, given the fact that the tokenization process requires hand crafted, man-made rules and knowledge about the constructs of the language which is something that can't be done automatically for now on all languages and specially the LREL ones. 

For model choice, our method is also model-generic meaning it can be used on any model given its non-intrusive property. But for the purpose of this research we will use the same architecture as ULMFiT, where we use a three-layer LSTM model that was pre-trained on wikitext-103 dataset \cite{DBLP:journals/corr/MerityXBS16} (which is a collection of over 100 million tokens extracted from the set of verified Good and Featured articles on Wikipedia). The later model is then fine-tuned on the training dataset in an unsupervised fashion and finally we add a 2 layer fully connected neural network as the classifier which is carefully fine-tuned on the training dataset using gradual unfreezing of layers\cite{DBLP:journals/corr/abs-1903-05987}. 

The Final part, is to introduce a new loss which operates on the final layers of the model. They act as enforcement for the model to dismiss the noise in the data and only extract features useful for classification. The new loss is inspired by Linear Discriminant Analysis (LDA) to capture the separability (based on the assumption that samples which are closer to each other in the latent space are classified similarly), Ficher defined an optimization function to maximize the between-class variability and minimize the within-class variability regardless of the source of the sample. the separability loss is defined as follows:

$$
\mathcal{L}_{sep}(W^E) =  \bigg(\frac{
\sum_{i \in Y} \sum_{z_{ij} \in Z_i} d(z_{ij}, \mu_i)
}{
\sum_{i \in Y} d(\mu_i, \mu)
}\bigg) \times \lambda_{BF}$$

$$
\lambda_{BF} = \frac{\min_{i} |Y^t_i|}{\max_{i} |Y^t_i|} \nonumber
$$

where $Z_i$ is the set of latent variables that belongs to class $i$ and it can be expressed as: $$Z_i = \cup_{s=1}^{N}Z_i^s \cup_{t=1}^{M} Z_i^t$$ which is the union of the sets of latent representation of all language domains that have the same label $i$. 
$\mu_i$ is the mean of the latent representations that has label $i$, so it can be expressed as $\mu_i = mean(Z_i)$, while $\mu$ is the mean of all the latent representations $\mu = mean(Z)$, the calculations of such loss can be hard so for simplicity we use this loss as batch based and in order to mitigate the drain of information we used a balancing factor $\lambda_{BF}$ which proved to be useful in other applications of the separability loss \cite{Bekkouch_2019}. $d(.,.)$ is the cosine dissimilarity which is used as a measure of distance between the samples. This loss allows us to increase the separability of the latent space according to the labels, regardless of its domain (language), which in return helps the classifier to generalize and provide more resilient and higher performances. 

\subsection{Dataset}
\subsubsection{Source Domain Dataset}
The Source domain dataset which is the training set of our model, used to predict tweets stances on unseen domain is a dataset for English Tweets. We use Semantic Evaluation competition train dataset. The data has 4163 samples with three features being ID, Target, tweet and Stance as the response variable.
Each tweet in the dataset has a target and can classified into stance class. The are five targets and three stances classes. The summary of the SemEval-2016 data set is presented in Figure \ref{figure:data_dist}

\subsubsection{Target Domain Dataset}
The target domain dataset language is Zulu. The Zulu language is a tonal language which also features click consonants. In comparison to English, in these clique consonants' noises are conveyed in literature with the word `tut!'. The Standard Zulu that is taught in schools is also referred to as `deep Zulu', as it uses many older Zulu words and phrases, and is altogether a purer form of the language than many of the dialects that are used in common speech. The Zulu language relies heavily on tone to convey meaning; but when the language itself is written down, often no tones are conveyed in the writing. This means that the speaker must have a good understanding of spoken Zulu before being able to read the Zulu language fluently, which is unusual among languages. All these interesting aspects and facts about our chosen target language makes it a perfect choice for Domain Generalization tasks since there exist a wide domain gap between itself and the English language (which is our source domain).
For our approach, we randomly sampled 1343 tweets from Semantic Evaluation competition test dataset (SemEval-2016) \cite{mohammad2016semeval}. We translated the tweets to Zulu language with the help of the Google Translate API, together with a native Zulu language speaker, to try to minimize the grammatical errors from the google translator. Examples of tweets with Zulu translation:

\begin{center}
\textbf{Atheism : AGAINST} \\
\textbf{English :} The humble trust in God: 'Whoever leans on, trusts in, and is confident in the Lord  happy, blessed, and fortunate is he' \\
\textbf{Zulu :} Abathobekile bathembela kuNkulunkulu: 'Noma ngubani oncika, athembela kuye, futhi othembela eNkosini uyajabula, ubusisiwe, futhi unenhlanhla'
\end{center}

\begin{center}
\textbf{Legalization of Abortion : FAVOR}  \\
\textbf{English :} Would you rather have women taking dangerous concoctions to induce abortions or know they are getting a safe \& legal one? \\
\textbf{Zulu :} Ngabe ungathanda ukuthi abesifazane bathathe imiqondo eyingozi ukukhipha izisu noma wazi ukuthi bathola ephephile futhi esemthethweni ?
\end{center}

\begin{center}
\textbf{Feminist Movement : NONE} \\
\textbf{English :} Some men do not deserve to be called gentlemen \\
\textbf{Zulu :} Amanye amadoda awakufanele ukubizwa ngokuthi ngamanenekazi
\end{center}

\begin{figure}[!ht] 
\centering
  \begin{subfigure}{\textwidth}
    \centering\includegraphics[width=\textwidth]{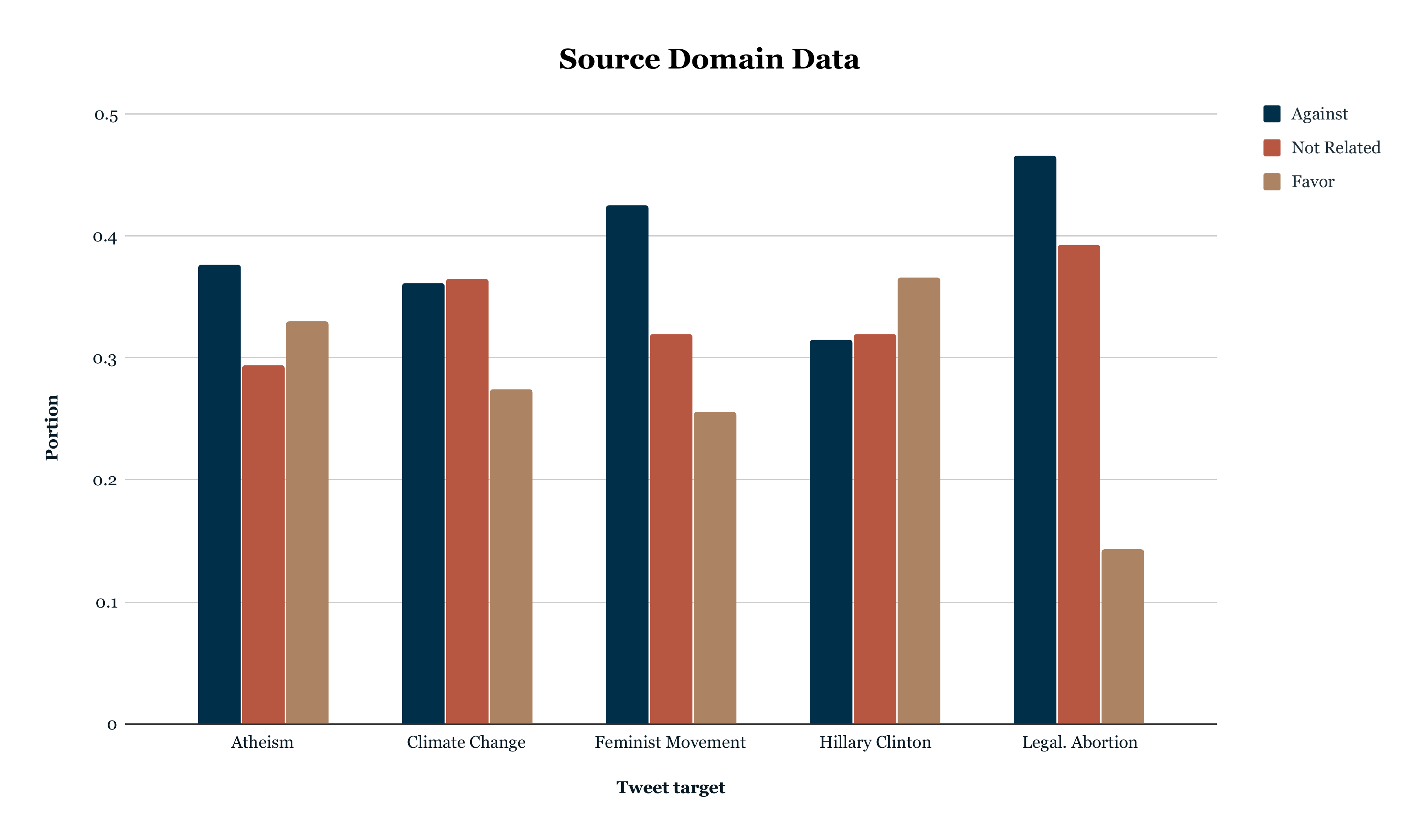}
    \caption{Source Domain: overall balanced w/ less data in Favor of Legal abortion.}
  \end{subfigure}
  \begin{subfigure}{\textwidth}
    \centering\includegraphics[width=\textwidth]{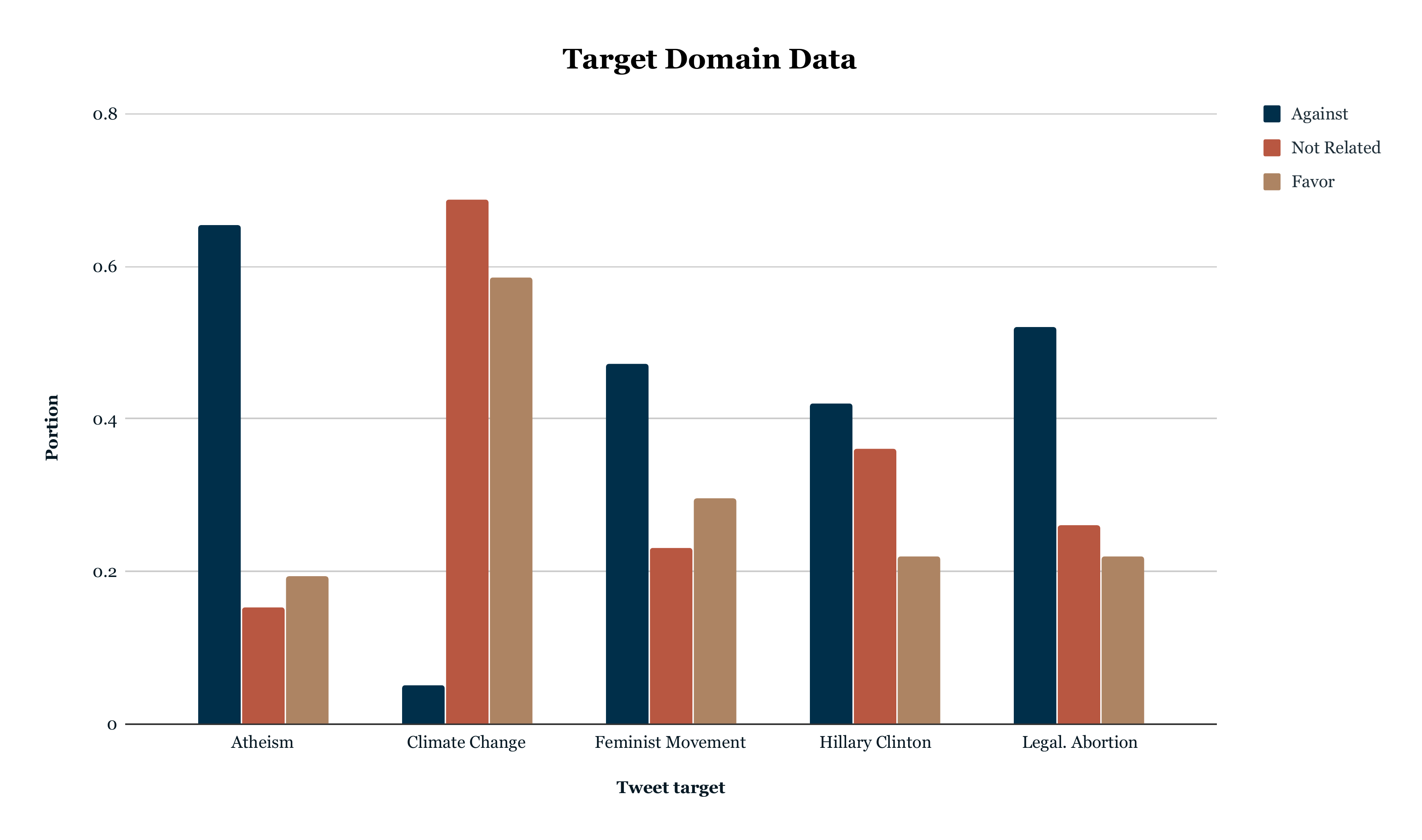}
    \caption{Target Domain: not as well balanced as the source dataset, with a big lack of data for 'against' in the 'Climate Change' topic and for both 'not-related' and 'in-favor' of the 'Atheism' topic.}
    \label{figure:after_DA}
  \end{subfigure}
  \caption{Dataset comparison between the source and the target dataset.}
  \label{figure:data_dist}
\end{figure}

The Stances and Targets distribution in our final target domain data is in Figure \ref{figure:data_dist}

\begin{table*}[!htbp]
\begin{center}
\resizebox{\textwidth}{!}{%
\begin{tabular}{|ccccc|cccc|}
\hline
Tested On                                     & \multicolumn{4}{c|}{\textbf{English}}                                                                                  & \multicolumn{4}{c|}{\textbf{Zulu}}                                                                                     \\ \cline{1-1}
\multicolumn{1}{|c|}{Trained On}              & \multicolumn{1}{c|}{F1-score} & \multicolumn{1}{c|}{Accuracy} & \multicolumn{1}{c|}{FAVOR-F1-score} & AGAINST-F1-score & \multicolumn{1}{c|}{F1-score} & \multicolumn{1}{c|}{Accuracy} & \multicolumn{1}{c|}{FAVOR-F1-score} & AGAINST-F1-score \\ \cline{2-9} 
\multicolumn{1}{|c|}{Zulu-Only (DLB)}            & -                            & -                            & -                                  & -               & 0.3942                            &  0.4861                            & 0.1929                                  & 0.4259               \\ \cline{2-9} 
\multicolumn{1}{|c|}{English-Only (DUB/DLB)}            & 0.5792                            & /                            & 0.4476                                  & 0.6908               & 0.4906                            & 0.5258                            & 0.3749                                  & 0.5742               \\ \cline{2-9} 
\multicolumn{1}{|c|}{Randomized-English-1}    & 0.5686                            & /                            & 0.4591                                  & 0.6951               & 0.5061                            & 0.5386                            & 0.3932                                  & 0.5987               \\ \cline{2-9} 
\multicolumn{1}{|c|}{Randomized-English-2}    & 0.5993                            & /                            & 0.4626                                  & 0.6999               & 0.5112                            & 0.5443                            & 0.3923                                  & 0.6147               \\ \cline{2-9} 
\multicolumn{1}{|c|}{Randomized-English-3}    & 0.6070                            & /                            & 0.4658                                  & 0.7083               & 0.5087                            & 0.5512                            & 0.3852                                  & 0.6090               \\ \cline{2-9} 
\multicolumn{1}{|c|}{Randomized-English-4}    & 0.6125                            & /                            & 0.4656                                  & \textbf{0.7243}               & 0.5186                            & \textbf{0.5635}                            & 0.4038                                  & 0.6204               \\ \cline{2-9} 
\multicolumn{1}{|c|}{Randomized-English-5}    & \textbf{0.6296}                            & /                            & \textbf{0.4710}                                  & 0.7201               & 0.5293                            & 0.548                            & 0.4014                                  & 0.6286               \\ \cline{2-9} 
\multicolumn{1}{|c|}{Randomized-English-Zulu} & 0.5690                            & /                            & 0.4586                                  & 0.7083               & \textbf{0.5493}                            & 0.5596                            & \textbf{0.4228}                                  & \textbf{0.6423}              \\ \hline
\end{tabular}%
}
\caption{Evaluation of Domain Randomization, We denote a randomized dataset with index i as the degree of randomization. We use the / symbol to denote that this result is not reported given that it is not accessible by the evaluation script of the original dataset. We also use - symbol to denote that this experiment isn't feasible. DUB and BLB are used to denote the baselines described in the baselines description \ref{subsec:base}}
\label{tab:domain_random}
\end{center}
\end{table*}

\subsection{Baselines}
\label{subsec:base}
For the purpose of evaluating our method, we will compare it's results against multiple baselines that represent an estimation of a lower-bound and an upper-bound of the performance.
\subsubsection{Lower Bound Baselines}
Our method should perform better than a model that was directly trained on English dataset and tested on a Zulu dataset (translated to English) and it should outperform a model trained only on the Zulu dataset and tested on it, given that we don't have enough data to fully train a model. that's why we will be using such models as our lower bound. These model are denoted as DLB (Direct Lower Bound).
\subsubsection{Upper Bound Baselines}
At the same time, Our model should not be able to exceed the performance of a model trained and tested on the same distribution meaning trained and tested on English without any translation from other languages. This model is denoted as DUB (Direct Upper Bound).

\section{Evaluation and Results}

We compare our model with several baselines on for Stance Detection and transfer learning. using the four metrics used for the stance detection challenge. 

\subsection{Domain Randomization}
We evaluate our stance detection model using Domain Randomization where we use the English dataset for Tweet Stance Detection, and test it on both its testing data and our Zulu dataset for Tweets Stance Detection. Here we notice that the model generalizes to itself even better the more randomization we add and the same for the Zulu dataset, as the results in Table~\ref{tab:domain_random} show. Results are reported with macro-F1-score and accuracy along with Favor-F1-score and Against-F1-score because the main challenge reported these metrics.

The different experiments were conducted all on the same dataset by adding different translated versions on top of it. The results reported are the mean 5 random runs of the models, where we noticed that on some experiments the results on the English dataset drops because the quality of the data changes drastically when the target language translation quality is too poor. 

We also noticed that by removing @ and \# and some other symbols, the accuracy increases and becomes more stable over several re-runs. It is also worth mentioning that the best results on the Zulu dataset: \textbf{F1-score = 0.5634} were achieved on Domain Randomization towards: English(original dataset), Zulu, Xhosa, Shona and Afrikaans languages \cite{niesler2005phonetic}, although we have no way of defining the reason for this increase it is most likely that it was achieved due to the similarities of the languages since they are all African languages even though Afrikaans is very different from Zulu, and due to the homogeneous performance of google Translate on these languages \cite{bourquin1951click}.

Some of the draw backs of the Domain Randomization technique is that it takes up to 100 times more time per epoch to train the model specially in the first unsupervised part and requires 3 to 5 times more epochs to converge(so that the accuracy and loss aren't changing a lot), plus the translation over-head which can take up to hours and can be faced by blocking IP-address by the google API. Another potential issue is that after a certain degree of randomization ($n_R = 16$ in our case) the models performance drops drastically even on the training data.

\subsection{Domain Adaptation}
We implemented a supervised domain adaptation scenario where we train on both English and Zulu datasets both with access to the labels. We used 70\% of the Zulu dataset for training and 30\% for testing. This is the only experiment where we used the Zulu dataset as part of the training set, unlike the Randomized-English-Zulu experiment where we only used Zulu as an intermediate language for the randomization process.

\begin{table}
\centering
\resizebox{8cm}{!}{%
\begin{tabular}{l|l|l|l|l|}
\cline{2-5}
 & F1-score & Accuracy & FA-F1-score & AG-F1-score \\ \hline
\multicolumn{1}{|l|}{Eng-Only} & 0.4906 & 0.5258 & 0.3749 & 0.5742 \\ \hline
\multicolumn{1}{|l|}{Eng-Zulu} & \textbf{0.5673} & \textbf{0.5448} & \textbf{0.4434} & \textbf{0.7469} \\ \hline
\end{tabular}
}
\caption{Domain Adaptation results on the Zulu-30 Test dataset.}
\label{tab:domain_adapt}
\end{table}

The results on table \ref{tab:domain_adapt} are the output of k-fold cross validation with k set to 10. The Domain Adaptation results show that there is definitely a visible increase in the performance on the Zulu test dataset (referred to as Zulu-30), it even outperforms the best domain randomization models evaluated in Table \ref{tab:domain_random}. 

\section{Conclusions}
We have proposed a non-intrusive method for improving results on Less-Resourced / Endangered Languages by leveraging low quality data from English datasets, using Zulu as a demonstration case. We also provide a new dataset for Stance Detection on Zulu. Our method was able to effectively transferring knowledge between different languages. In future work, we aim to improve the technique by adding more intrusive techniques like distribution matching between Source and Target domains in the latent space by reducing KL divergence.

\section*{Acknowledgments}

This research was supported by the Independent Danish Research Fund through the Verif-AI project grant.

\bibliographystyle{spbasic}
\bibliography{references.bib}

\begin{thebibliography}{33}
\providecommand{\natexlab}[1]{#1}
\providecommand{\url}[1]{{#1}}
\providecommand{\urlprefix}{URL }
\expandafter\ifx\csname urlstyle\endcsname\relax
  \providecommand{\doi}[1]{DOI~\discretionary{}{}{}#1}\else
  \providecommand{\doi}{DOI~\discretionary{}{}{}\begingroup
  \urlstyle{rm}\Url}\fi
\providecommand{\eprint}[2][]{\url{#2}}

\bibitem[{Allah and Boulaknadel(2012)}]{allah2012toward}
Allah FA, Boulaknadel S (2012) Toward computational processing of less
  resourced languages: Primarily experiments for moroccan amazigh language.
  Text Mining Rijeka: InTech pp 197--218

\bibitem[{Augenstein et~al(2016)Augenstein, Rockt{\"{a}}schel, Vlachos, and
  Bontcheva}]{DBLP:journals/corr/AugensteinRVB16}
Augenstein I, Rockt{\"{a}}schel T, Vlachos A, Bontcheva K (2016) Stance
  detection with bidirectional conditional encoding. In: Proceedings of the
  2016 Conference on Empirical Methods in Natural Language Processing, vol
  abs/1606.05464, \urlprefix\url{http://arxiv.org/abs/1606.05464},
  \eprint{1606.05464}

\bibitem[{Bekkouch et~al(2019)Bekkouch, Youssry, Gafarov, Khan, and
  Khattak}]{Bekkouch_2019}
Bekkouch IEI, Youssry Y, Gafarov R, Khan A, Khattak AM (2019) Triplet loss
  network for unsupervised domain adaptation. Algorithms 12(5):96,
  \doi{10.3390/a12050096}, \urlprefix\url{http://dx.doi.org/10.3390/a12050096}

\bibitem[{Besacier et~al(2014)Besacier, Barnard, Karpov, and
  Schultz}]{besacier2014automatic}
Besacier L, Barnard E, Karpov A, Schultz T (2014) Automatic speech recognition
  for under-resourced languages: A survey. Speech Communication 56:85--100

\bibitem[{Bourquin(1951)}]{bourquin1951click}
Bourquin W (1951) Click-words which xhosa, zulu and sotho have in common.
  African Studies 10(2):59--81

\bibitem[{Carlucci et~al(2019)Carlucci, D'Innocente, Bucci, Caputo, and
  Tommasi}]{DBLP:journals/corr/abs-1903-06864}
Carlucci FM, D'Innocente A, Bucci S, Caputo B, Tommasi T (2019) Domain
  generalization by solving jigsaw puzzles. CoRR abs/1903.06864,
  \urlprefix\url{http://arxiv.org/abs/1903.06864}, \eprint{1903.06864}

\bibitem[{Cope(1966)}]{cope1966Zulu}
Cope AT (1966) Zulu phonology, tonology and tonal grammar. PhD thesis,
  University of Durban

\bibitem[{Derczynski et~al(2013)Derczynski, Maynard, Aswani, and
  Bontcheva}]{derczynski2013microblog}
Derczynski L, Maynard D, Aswani N, Bontcheva K (2013) Microblog-genre noise and
  impact on semantic annotation accuracy. In: Proceedings of the 24th ACM
  Conference on Hypertext and Social Media, ACM, pp 21--30

\bibitem[{Devlin et~al(2018)Devlin, Chang, Lee, and Toutanova}]{devlin2018BERT}
Devlin J, Chang MW, Lee K, Toutanova K (2018) Bert: Pre-training of deep
  bidirectional transformers for language understanding. \eprint{1810.04805}

\bibitem[{Dungs et~al(2018)Dungs, Aker, Fuhr, and Bontcheva}]{dungs2018can}
Dungs S, Aker A, Fuhr N, Bontcheva K (2018) Can rumour stance alone predict
  veracity? In: Proceedings of the 27th International Conference on
  Computational Linguistics, pp 3360--3370

\bibitem[{Ferraro et~al(2013)Ferraro, Daum{\'e}~III, DuVall, Chapman, Harkema,
  and Haug}]{ferraro2013improving}
Ferraro JP, Daum{\'e}~III H, DuVall SL, Chapman WW, Harkema H, Haug PJ (2013)
  Improving performance of natural language processing part-of-speech tagging
  on clinical narratives through domain adaptation. Journal of the American
  Medical Informatics Association 20(5):931--939

\bibitem[{Goodfellow(2016)}]{goodfellow2016nips}
Goodfellow I (2016) Nips 2016 tutorial: Generative adversarial networks. arXiv
  preprint arXiv:170100160

\bibitem[{Gorrell et~al(2019)Gorrell, Kochkina, Liakata, Aker, Zubiaga,
  Bontcheva, and Derczynski}]{gorrell2019semeval}
Gorrell G, Kochkina E, Liakata M, Aker A, Zubiaga A, Bontcheva K, Derczynski L
  (2019) Semeval-2019 task 7: Rumoureval, determining rumour veracity and
  support for rumours. In: Proceedings of the 13th International Workshop on
  Semantic Evaluation, pp 845--854

\bibitem[{Hercig et~al(2018)Hercig, Krejzl, and Kr{\'a}l}]{hercig2018stance}
Hercig T, Krejzl P, Kr{\'a}l P (2018) Stance and sentiment in czech.
  Computaci{\'o}n y Sistemas 22(3)

\bibitem[{Howard and Ruder(2018)}]{DBLP:journals/corr/abs-1801-06146}
Howard J, Ruder S (2018) Fine-tuned language models for text classification.
  CoRR abs/1801.06146, \urlprefix\url{http://arxiv.org/abs/1801.06146},
  \eprint{1801.06146}

\bibitem[{{Hu} et~al(2018){Hu}, {Kan}, {Shan}, and {Chen}}]{8578260}
{Hu} L, {Kan} M, {Shan} S, {Chen} X (2018) Duplex generative adversarial
  network for unsupervised domain adaptation. In: 2018 IEEE/CVF Conference on
  Computer Vision and Pattern Recognition, pp 1498--1507,
  \doi{10.1109/CVPR.2018.00162}

\bibitem[{Kochkina et~al(2017)Kochkina, Liakata, and
  Augenstein}]{DBLP:journals/corr/KochkinaLA17}
Kochkina E, Liakata M, Augenstein I (2017) Proceedings of the 11th
  international workshop on semantic evaluation (semeval-2017). In: CoRR, vol
  abs/1704.07221, \urlprefix\url{http://arxiv.org/abs/1704.07221},
  \eprint{1704.07221}

\bibitem[{Kotu and Deshpande(2015)}]{KOTU201517}
Kotu V, Deshpande B (2015) Chapter 2 - data mining process. In: Kotu V,
  Deshpande B (eds) Predictive Analytics and Data Mining, Morgan Kaufmann,
  Boston, pp 17 -- 36,
  \doi{https://doi.org/10.1016/B978-0-12-801460-8.00002-1},
  \urlprefix\url{http://www.sciencedirect.com/science/article/pii/B9780128014608000021}

\bibitem[{Kotu and Deshpande(2019)}]{KOTU201919}
Kotu V, Deshpande B (2019) Chapter 2 - data science process. In: Kotu V,
  Deshpande B (eds) Data Science (Second Edition), second edition edn, Morgan
  Kaufmann, pp 19 -- 37,
  \doi{https://doi.org/10.1016/B978-0-12-814761-0.00002-2},
  \urlprefix\url{http://www.sciencedirect.com/science/article/pii/B9780128147610000022}

\bibitem[{K{\"u}{\c{c}}{\"u}k(2017)}]{kuccuk2017stance}
K{\"u}{\c{c}}{\"u}k D (2017) Stance detection in turkish tweets. arXiv preprint
  arXiv:170606894

\bibitem[{Li et~al(2017)Li, Yang, Song, and
  Hospedales}]{DBLP:journals/corr/abs-1710-03077}
Li D, Yang Y, Song Y, Hospedales TM (2017) Deeper, broader and artier domain
  generalization. CoRR abs/1710.03077,
  \urlprefix\url{http://arxiv.org/abs/1710.03077}, \eprint{1710.03077}

\bibitem[{Lillie et~al(2019)Lillie, Middelboe, and
  Derczynski}]{lillie2019joint}
Lillie AE, Middelboe ER, Derczynski L (2019) Joint rumour stance and veracity
  prediction. In: Proceedings of the 22nd Nordic Conference on Computional
  Linguistics (NoDaLiDa), pp 208--221

\bibitem[{Mahsut et~al(2001)Mahsut, Ogawa, Sugino, and
  Inagaki}]{mahsut2001utilizing}
Mahsut M, Ogawa Y, Sugino K, Inagaki Y (2001) Utilizing agglutinative features
  in japanese-uighur machine translation. In: Proceedings of MT Summit, vol~8,
  pp 217--222

\bibitem[{Merity et~al(2016)Merity, Xiong, Bradbury, and
  Socher}]{DBLP:journals/corr/MerityXBS16}
Merity S, Xiong C, Bradbury J, Socher R (2016) Pointer sentinel mixture models.
  CoRR abs/1609.07843, \urlprefix\url{http://arxiv.org/abs/1609.07843},
  \eprint{1609.07843}

\bibitem[{Mohammad et~al(2016)Mohammad, Kiritchenko, Sobhani, Zhu, and
  Cherry}]{mohammad2016semeval}
Mohammad S, Kiritchenko S, Sobhani P, Zhu X, Cherry C (2016) Semeval-2016 task
  6: Detecting stance in tweets. In: Proceedings of the 10th International
  Workshop on Semantic Evaluation (SemEval-2016), pp 31--41

\bibitem[{Niesler et~al(2005)Niesler, Louw, and Roux}]{niesler2005phonetic}
Niesler T, Louw P, Roux J (2005) Phonetic analysis of afrikaans, english, xhosa
  and zulu using south african speech databases. Southern African Linguistics
  and Applied Language Studies 23(4):459--474

\bibitem[{Nisbet et~al(2009)Nisbet, Elder, and Miner}]{NISBET2009285}
Nisbet R, Elder J, Miner G (2009) Chapter 13 - model evaluation and
  enhancement. In: Nisbet R, Elder J, Miner G (eds) Handbook of Statistical
  Analysis and Data Mining Applications, Academic Press, Boston, pp 285 -- 312,
  \doi{https://doi.org/10.1016/B978-0-12-374765-5.00013-9},
  \urlprefix\url{http://www.sciencedirect.com/science/article/pii/B9780123747655000139}

\bibitem[{Peters et~al(2019)Peters, Ruder, and
  Smith}]{DBLP:journals/corr/abs-1903-05987}
Peters ME, Ruder S, Smith NA (2019) To tune or not to tune? adapting pretrained
  representations to diverse tasks. CoRR abs/1903.05987,
  \urlprefix\url{http://arxiv.org/abs/1903.05987}, \eprint{1903.05987}

\bibitem[{Pires et~al(2019)Pires, Schlinger, and
  Garrette}]{DBLP:journals/corr/abs-1906-01502}
Pires T, Schlinger E, Garrette D (2019) How multilingual is multilingual bert?
  CoRR abs/1906.01502, \urlprefix\url{http://arxiv.org/abs/1906.01502},
  \eprint{1906.01502}

\bibitem[{Qazvinian et~al(2011)Qazvinian, Rosengren, Radev, and
  Mei}]{qazvinian2011rumor}
Qazvinian V, Rosengren E, Radev DR, Mei Q (2011) Rumor has it: Identifying
  misinformation in microblogs. In: Proceedings of the conference on empirical
  methods in natural language processing, Association for Computational
  Linguistics, pp 1589--1599

\bibitem[{Tobin et~al(2017)Tobin, Fong, Ray, Schneider, Zaremba, and
  Abbeel}]{DBLP:journals/corr/TobinFRSZA17}
Tobin J, Fong R, Ray A, Schneider J, Zaremba W, Abbeel P (2017) Domain
  randomization for transferring deep neural networks from simulation to the
  real world. CoRR abs/1703.06907,
  \urlprefix\url{http://arxiv.org/abs/1703.06907}, \eprint{1703.06907}

\bibitem[{Weng(2019)}]{weng2019DR}
Weng L (2019) Domain randomization for sim2real transfer.
  lilianwenggithubio/lil-log
  \urlprefix\url{http://lilianweng.github.io/lil-log/2019/05/04/domain-randomization.html}

\bibitem[{{Zhou} et~al(2019){Zhou}, {Lin}, {Tan}, and {Liu}}]{8851965}
{Zhou} S, {Lin} J, {Tan} L, {Liu} X (2019) Condensed convolution neural network
  by attention over self-attention for stance detection in twitter. In: 2019
  International Joint Conference on Neural Networks (IJCNN), pp 1--8,
  \doi{10.1109/IJCNN.2019.8851965}

\end{thebibliography}

\end{document}